# Distributed Reinforcement Learning for Cooperative Multi-Robot Object Manipulation

## Extended Abstract

Guohui Ding*
University of Colorado Boulder

Joewie J. Koh*
University of Colorado Boulder

Kelly Merckaert†
Vrije Universiteit Brussel

Bram Vanderborght†
Vrije Universiteit Brussel

Marco M. Nicotra
University of Colorado Boulder

Christoffer Heckman
University of Colorado Boulder

Alessandro Roncone
University of Colorado Boulder

Lijun Chen
University of Colorado Boulder

## ABSTRACT

We consider solving a cooperative multi-robot object manipulation task using reinforcement learning (RL). We propose two distributed multi-agent RL approaches: distributed approximate RL (DA-RL), where each agent applies Q-learning with individual reward functions; and game-theoretic RL (GT-RL), where the agents update their Q-values based on the Nash equilibrium of a bimatrix Q-value game. We validate the proposed approaches in the setting of cooperative object manipulation with two simulated robot arms. Although we focus on a small system of two agents in this paper, both DA-RL and GT-RL apply to general multi-agent systems, and are expected to scale well to large systems.

## 1 INTRODUCTION

Deep reinforcement learning (RL) has recently been successfully applied to various robot control problems [9], mostly in single-agent settings. Generalization to multi-agent settings such as multi-robot cooperation is extremely challenging, and requires scalable solutions for managing the large number of degrees of freedom (DoFs), heterogeneous physical constraints, and possibly partial or asymmetric observations at different robots.

Parallel implementations of single-agent RL on different agents scale well to large multi-agent systems, but suffer from issues such as learning instability. This is due to non-stationarity of the environment that each agent faces [4]. To ensure good performance, the agents should be jointly trained [1, 3, 8, 13], in a distributed manner. In this paper, we propose two distributed multi-agent RL approaches. In distributed approximate RL (DA-RL), each agent applies Q-learning with individual reward functions being coupled and aligned to the goal of the cooperative task. In game-theoretic RL (GT-RL), the agents update their Q-values based on the Nash equilibrium of the bimatrix game of estimated Q-values.

*These authors contributed equally to this work. Please direct all correspondence to Guohui Ding (guohui.ding@colorado.edu).
†These authors are also affiliated with Flanders Make.

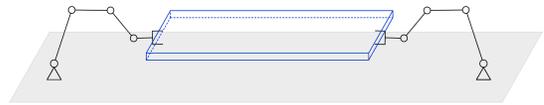

Figure 1: Each RL agent is responsible for one manipulator.

We validate our methods in the setting of cooperative object manipulation (see Figure 1) with two Sawyer robotic manipulators in the Gazebo simulator [7], but the methods are applicable to other combinations of manipulators as well. In our scenario, the state of robot $k$ at time $t$ is given by $s_t = (q_t^k, p_t^k)_{k \in \mathcal{N}}$, where $q_t^k$ denotes the joint angles and $p_t^k$ is the global coordinate of the robot's end effector. The state-control dynamics for the $i$-th joint is then $q_{t+1}^{k,i} = q_t^{k,i} + \Delta a^{k,i} \Delta t + \epsilon_t^{k,i}$, where $\Delta a^{k,i}$ is the joint angle increment.

## 2 DISTRIBUTED RL

The system for the multi-agent RL can be defined as the tuple $\mathcal{M} = \{\mathcal{N}, S, (A^k)_{k \in \mathcal{N}}, P, (r^k)_{k \in \mathcal{N}}, (\pi^k)_{k \in \mathcal{N}}, V^k(\cdot)\}$, with the set $\mathcal{N} = \{1, \ldots, n\}$ of agents, the set $S$ of states, the set $A^k$ of actions available to the agent $k$, the state transition function $P \colon S \times A^1 \times \cdots \times A^n \to PD(S)$, the reward function $r^k \colon S \times A^1 \times \cdots \times A^n \to \mathbb{R}$, and the policy $\pi^k \colon S \to PD(A^k)$ for the agent $k$. Each agent $k \in \mathcal{N}$ seeks to maximize its accumulated reward with the starting state $s$ at time $t$: $V^k(s) = \mathbf{E}[\sum_{i=0}^{T-t} \gamma^i r_{t+i}^k \mid s_t = s, (\pi^k)_{k \in \mathcal{N}}]$.

We consider a setting where the agents share the common observed state $s_t$, but try to learn a deterministic policy $\pi^k \colon S \to A^k$ individually. In such a setting, the key to distributed RL for cooperative tasks is the proper engineering of individual reward functions $r^k$ that capture the common goal of the task while respecting the distributed structure of the system. Specific to the bi-robot object manipulation problem, we first identify the key constituents of the reward for the task: i) those that capture the object displacement from target, respectively $r^{g_1} = -d(p_{t+1}^1, p_{target}^1)$ and $r^{g2} = -d(p^2_{t+1}, p^2_{target})$ for the two robots; and ii) that which captures the object posture deviation, $r^{g3} = -a(p^1_{t+1}, p^2_{t+1})$. Here $d(p, p') = \|p - p'\|_1$ characterizes the distance between $p$ and $p'$, and $a(p^1, p^2)$ is the absolute angle between the vector $(p^1 - p^2)$ and the target $(p^1_{target} - p^2_{target})$. We then present two ways of distributing the afore three constituents to the two robots:

**RS-1:** $\begin{cases} r^1(s_t, a^1_t) = -d(p^1_{t+1}, p^1_{target}) - \kappa_1 a(p^1_{t+1}, p^2_{t+1}), \\ r^2(s_t, a^2_t) = -d(p^2_{t+1}, p^2_{target}) - \kappa_2 a(p^1_{t+1}, p^2_{t+1}); \end{cases}$ (1)

**RS-2:** $\begin{cases} r^1(s_t, a^1_t) = -d(p^1_{t+1}, p^1_{target}) - d(p^2_{t+1}, p^2_{target}), \\ r^2(s_t, a^2_t) = -\kappa a(p^1_{t+1}, p^2_{t+1}). \end{cases}$ (2)

In **RS-1**, each robot is concerned with both its end effector displacement to the target and the object posture deviation. In **RS-2**, one robot is concerned with the object displacement to the target, while the other is concerned with the object posture deviation. Here, $\kappa_1 > 0$, $\kappa_2 > 0$, and $\kappa > 0$ are parameters used to strike different tradeoffs between different constituents, where $\kappa_1 + \kappa_2 = \kappa$.

Given a multi-agent system, it is of paramount importance to identify a reasonable outcome from the multi-agent interaction and determine how to achieve that outcome. This can be approached from two complementary perspectives. One perspective starts with a specification of individual agent behavior (e.g., learning algorithm in our problem), and then studies how the resulting system performs. The other perspective starts with a specification of reasonable outcome such as the Nash equilibrium [10], and then studies how to achieve that outcome. Next, we present both perspectives.

### 2.1 Distributed Approximate RL (DA-RL)

Similarly to independent RL methods [11, 14], each agent (i.e., robot) $k$ applies the single-agent Q-learning with individual reward $r^k(s_t, a^k_t)$ [12, 16]. Like those methods, this distributed RL approach does not have a convergence guarantee. However, notice that the sum of individual agent rewards $r(s_t, a^1_t, a^2_t) = r^1(s_t, a^1_t) + r^2(s_t, a^2_t)$ is a well-defined systemwide reward for the whole multi-agent system. Our approach can be seen as a distributed approximation of the centralized RL with reward function $r(s_t, a^1_t, a^2_t)$, and hence termed distributed approximate RL (DA-RL).

### 2.2 Game-Theoretic RL (GT-RL)

We now take the second perspective, and use the Nash equilibrium as the solution concept for the desired outcome from the multi-agent interaction. Specifically, we consider the general-sum Markov game formulation for our bi-robot object manipulation problem, with $\mathcal{N} = \{1, 2\}$. The Markov game (a.k.a., stochastic game) $\mathcal{G}$ includes the same components as $\mathcal{M}$ [10]. However, instead of direct optimization of $V^k(\cdot)$, agents aim to reach a Nash equilibrium $(\pi^{1^*}, \pi^{2^*})$ of $\mathcal{G}$ such that $\forall s \in S$,

$$V^1(s \mid \pi^{1^*}, \pi^{2^*}) \geq V^1(s \mid \pi^1, \pi^{2^*}), \quad \forall \pi^1; \quad (3)$$

$$V^2(s \mid \pi^{1^*}, \pi^{2^*}) \geq V^2(s \mid \pi^{1^*}, \pi^2), \quad \forall \pi^2. \quad (4)$$

At the Nash equilibrium, no agent has incentive to change its policy given that the other agent takes the equilibrium policy. Also note that, in our problem, we consider only the stationary policies $\pi^k = (\pi^k(s^1), \pi^k(s^2), \dots)$, and hence the existence of the Nash equilibrium is guaranteed by Theorem 4.6.4 in [2].

Having specified the Nash equilibrium as a desired outcome, we leverage Nash-Q learning [6, 10, 15] and deep Q-networks to design a distributed RL scheme, termed game-theoretic RL (GT-RL), to achieve the equilibrium for the bi-robot object manipulation problem. The convergence to the Nash equilibrium guarantees the stability of the policies learned by each agent, effectively accounting for the non-stationary nature of multi-agent systems.

Specifically, at step $t$, agent $k$ observes state $s_t$ and takes action $a^k_t \sim \pi^k(\cdot \mid s_t)$ with the help of $Q^k_\pi(s_t, (a^1_t, a^2_t))$. Each step can be considered as a bimatrix game $\mathcal{G}^{bi}(s) = \{\mathcal{N}, (Q^k)_{k \in \mathcal{N}}, (\pi^k)_{k \in \mathcal{N}}, s\}$. The derived bimatrix equilibrium $(\mu^{1^*}, \mu^{2^*})$ satisfies:

$$\mu^{1^*} M^1 \mu^{2^*} \geq \mu^1 M^1 \mu^{2^*}, \quad \forall \mu^1 \in PD(A^1); \quad (5)$$

$$\mu^{1^*} M^2 \mu^{2^*} \geq \mu^{1^*} M^2 \mu^2, \quad \forall \mu^2 \in PD(A^2). \quad (6)$$

In fact, the whole Markov game can be decomposed as a sequence of bimatrix games $\mathcal{G} = \{\mathcal{G}^{bi}(s_t)\}_{t=1,2,\dots}$, and Theorem 3 in [5] ensures that the Nash equilibrium $(\mu^{1^*}, \mu^{2^*})$ of $\mathcal{G}^{bi}(s_t)$ is also part of the Nash equilibrium of $\mathcal{G}$. Hence, if we assume that the Nash equilibrium of $\mathcal{G}$ is defined by $\pi^{k^*} = \{\pi^{k^*}(\bar{s}_\ell)\}_{\ell=1,\dots,|S|}$, there exists one $(\pi^{1^*}(\bar{s}'), \pi^{2^*}(\bar{s}'))$ that is also the Nash equilibrium of $\mathcal{G}^{bi}(s_t)$ with $s_t = \bar{s}'$ and $\pi^k(s_t) = \mu^{k^*}$. The $Q^k_t$ update [5] is thus Eq. (7):

$$Q^k_{t+1}(s_t, (a^1, a^2)) = (1 - \alpha_t) Q^k_t(s_t, (a^1, a^2)) + \alpha_t [r^k_{t+1} + \gamma \pi^1(s_{t+1}) Q^k_t(s_{t+1}) \pi^2(s_{t+1})]. \quad (7)$$

## 3 RESULTS AND CONCLUSION

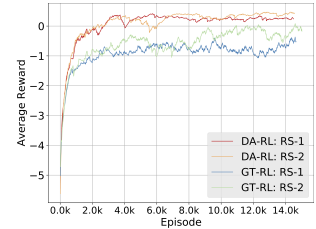 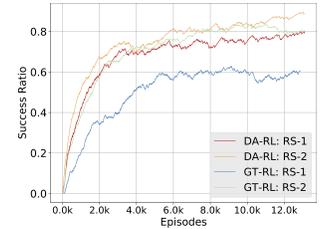

Figure 2: Learning curve.   Figure 3: Success ratio.

During training, both methods exhibit stable policy learning curves, as shown in Figure 2 where the y-axis value is the sum of average $r^1$ and $r^2$ at each episode. After 4,000 episodes of training, the two methods with different reward structures reach convergence and continue finetuning the learned polices. Figure 3 shows the success ratios for the two methods, averaged over 10 different random seeds. DA-RL attains a success ratio of 80% within 8,000 episodes. With GT-RL, RS-2 leads to better performance compared to RS-1. We conclude that DA-RL is more robust to the use of different reward structures, but GT-RL is more sensitive to choice of reward structure. Although we focus on a small system of two agents, both DA-RL and GT-RL apply to general multi-agent systems and are expected to scale well to large systems.

## ACKNOWLEDGMENTS

This material is based upon work supported by the National Science Foundation under Grant No. 1646556, and by the Flemish Government under the program Onderzoeksprogramma Artificiële Intelligentie (AI) Vlaanderen.